%% file: main.tex
\input{preamble_icves.tex}

\begin{document}
\icvesanonymousfalse
\icvesarxivtrue
\SetICVESBalanceAdjustment{-0.95em}
\AddICVESArxivHeader
\AddIEEECopyrightFooter
\input{content_icves.tex}
\end{document}

%% file: preamble_icves.tex
\documentclass[letterpaper, 10 pt, conference]{ieeeconf}  

\IEEEoverridecommandlockouts                              

\usepackage{cite}
\usepackage{amsmath,amssymb,amsfonts}
\usepackage{graphicx}
\usepackage{textcomp}
\usepackage{xcolor}
\usepackage{array}
\usepackage[caption=false,font=footnotesize]{subfig}
\usepackage{url}
\usepackage{tikz}
\usetikzlibrary{arrows.meta, positioning, calc, backgrounds, fit}
\usepackage{eso-pic}

\usepackage[hidelinks]{hyperref}

\newif\ificvesanonymous
\icvesanonymousfalse
\newif\ificvesarxiv
\icvesarxivfalse

\newlength{\icvesbalanceadjust}
\newcommand{\SetICVESBalanceAdjustment}[1]{%
\setlength{\icvesbalanceadjust}{#1}%
}
\newcommand{\ApplyICVESBalanceAdjustment}{%
\addtolength{\textheight}{\icvesbalanceadjust}%
}

\newcommand{\icvescitebagzel}{%
\ificvesanonymous
.
\else
  \space such as Bagzel \cite{pohl2026moda,pohl2026bag}.%
\fi
}

\newcommand{\icvestitle}{Safety-Oriented Trajectory Ranking from BEV Perception Outputs}

\newcommand{\icvesthanksanonymous}{%
\thanks{Corresponding author information is omitted for double-blind review.\protect\\
Code is available in an anonymized repository:
\protect\url{https://anonymous.4open.science/r/bev-trajectory-eval-6355}.}%
}

\newcommand{\icvesthanksfull}{%
\thanks{Corresponding author: Leon Pohl, \texttt{leon.pohl@unibw.de}.\protect\\
Code: \href{https://github.com/UniBwTAS/bev-trajectory-eval}{github.com/UniBwTAS/bev-trajectory-eval}.}%
}

\newcommand{\icvesauthorsanonymous}{%
\authorblockN{%
Anonymous Author\authorrefmark{1},
Anonymous Author\authorrefmark{2},
Anonymous Author\authorrefmark{3}, and
Anonymous Author\authorrefmark{1}\icvesthanksanonymous%
}
\authorblockA{\authorrefmark{1}
Anonymous Institution,
Anonymous City, Anonymous Country}
\authorblockA{\authorrefmark{2}
Anonymous Institution,
Anonymous City, Anonymous Country}
\authorblockA{\authorrefmark{3}
Anonymous Institution,
Anonymous City, Anonymous Country}%
}

\newcommand{\icvesauthorsfull}{%
\authorblockN{%
Leon Pohl\authorrefmark{1},
Martin Frauendorfer\authorrefmark{2},
Corinna Schmitt\authorrefmark{3}, and
Mirko Maehlisch\authorrefmark{1}\icvesthanksfull%
}
\authorblockA{\authorrefmark{1}
Institute for Autonomous Driving, University of the Bundeswehr Munich,
Neubiberg, Germany}
\authorblockA{\authorrefmark{2}
Ludwig-Maximilians-Universit\"at M\"unchen, Munich, Germany}
\authorblockA{\authorrefmark{3}
Secure Communication Systems, Research Institute CODE,
University of the Bundeswehr Munich, Neubiberg, Germany}%
}

\newcommand{\icvessetuptitle}{%
\title{\icvestitle}
\ificvesanonymous
  \author{\icvesauthorsanonymous}
\else
  \author{\icvesauthorsfull}
\fi
}

\newcommand{\AddICVESArxivHeader}{%
\AddToShipoutPictureFG*{%
  \AtPageUpperLeft{%
    \hspace{0.65in}%
    \raisebox{-0.43in}{%
      \parbox{7.2in}{%
        \scriptsize 2026 IEEE International Conference on Vehicular Electronics and Safety (ICVES 2026). Accepted Author Manuscript, October 2026.
      }%
    }%
  }%
}%
}

\newcommand{\AddIEEECopyrightFooter}{%
\AddToShipoutPictureFG*{%
  \AtPageLowerLeft{%
    \hspace{0.65in}%
    \raisebox{0.35in}{%
      \parbox{7.2in}{%
        \scriptsize
        \textcopyright{} 2026 IEEE. Personal use of this material is permitted.
        Permission from IEEE must be obtained for all other uses, in any current
        or future media, including reprinting/republishing this material for
        advertising or promotional purposes, creating new collective works, for
        resale or redistribution to servers or lists, or reuse of any copyrighted
        component of this work in other works.
      }%
    }%
  }%
}%
}

\def\BibTeX{{\rm B\kern-.05em{\sc i\kern-.025em b}\kern-.08em
    T\kern-.1667em\lower.7ex\hbox{E}\kern-.125emX}}

%% file: content_icves.tex
\bstctlcite{IEEEexample:BSTcontrol}

\icvessetuptitle
\maketitle

\begin{abstract}
Bird's-Eye-View (BEV) representations are increasingly used as compact inputs for autonomous driving perception and planning. However, lightweight and interpretable methods for evaluating candidate ego trajectories directly on BEV outputs remain useful for vehicular safety workflows, rapid prototyping, and human-in-the-loop validation. This paper presents a modular trajectory ranking pipeline that consumes BEVFusion object detections and map segmentations derived from nuScenes sensor data and ranks manually specified candidate trajectories according to safety-first criteria. The ranking prioritizes avoiding collisions and walkway violations, then smaller end distance to the target, and finally path efficiency. To study whether these rule-based rankings resemble human judgment, we created seven BEV traffic scenarios and compared the script output with preferences from a survey of 21 participants. Across 146 pairwise votes, 124 matched the trajectory ranked first by the evaluator, corresponding to an overall agreement of 84.9\%. The results indicate that simple BEV-centric safety heuristics can approximate human preferences in constrained trajectory choice tasks while remaining transparent enough for analysis, dataset curation, and autonomous driving studies.
\end{abstract}


\begin{figure}[!t]
\centering
\input{figs/teaser.tex}
\caption{Overview of the static BEV-based trajectory assessment setting. 
BEVFusion object detections and map segmentation outputs are mapped into a unified BEV scenario representation. Candidate trajectories are evaluated on the shared BEV grid using collision and walkway penalties, then ranked lexicographically by cumulative score, end distance to target, and path length.}
\label{fig:compact-bev-ranking}
\end{figure}
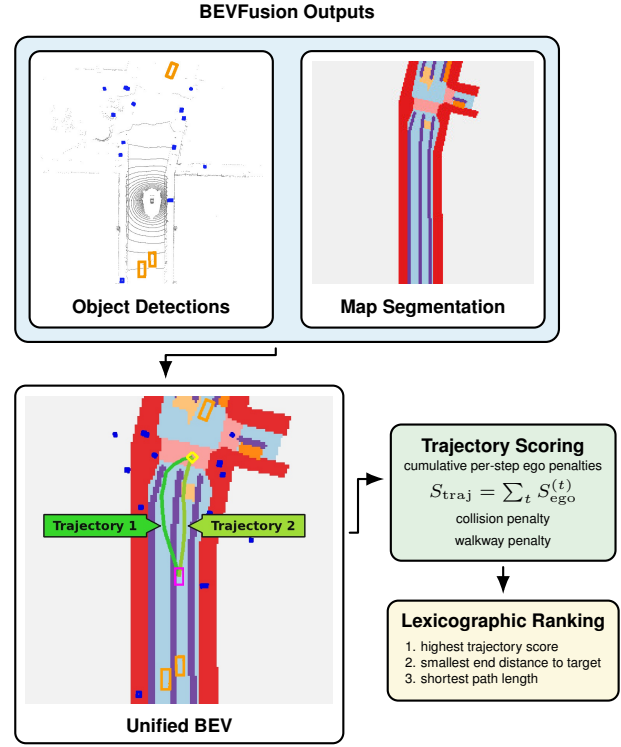

\section{Introduction}
Autonomous driving stacks rely on perception, prediction, planning, and decision-making modules that must operate safely in complex traffic scenes \cite{chen2024end,zhao2024aut}. Bird's-Eye-View (BEV) representations are attractive in this setting because they condense multi-sensor observations into a unified spatial format that is easier to use for downstream reasoning than raw camera or lidar streams alone \cite{philion2020lif,li2025bev,liu2023bev}.

As BEV perception increasingly serves as an interface between sensing and planning, it is important to understand whether its outputs provide sufficient semantic structure for safety-oriented trajectory assessment. This question is closely related to world-model-based reasoning, where compact scene representations support downstream prediction and decision-making \cite{ha2018rec,hafner2025traa}. Before adding learned temporal prediction or interaction modeling, we first examine how far static BEV scene information alone can support interpretable trajectory-cost reasoning.

We study the following question: \emph{To what extent do rule-based trajectory rankings generated from BEV perception outputs align with human trajectory preferences in safety-relevant driving scenarios?}

To answer this question, we evaluate candidate ego trajectories directly on BEVFusion object detections and map segmentations derived from nuScenes data. The resulting ranking uses transparent safety-first criteria, including object collisions, walkway violations, end distance to the target, and path length. We compare the induced rankings with human judgments in seven controlled scenario pairs.

Our contribution is threefold:
\begin{itemize}
\item We present a modular trajectory ranking pipeline that uses BEVFusion perception outputs for safety evaluation (cf. Fig.~\ref{fig:compact-bev-ranking}). 
\item We define an interpretable safety-first ranking rule over candidate ego trajectories using collision penalties, walkway penalties, end distance to the target, and path length.
\item We compare the resulting rankings with human judgments in seven scenario pairs and observe an overall agreement of 84.9\%.
\end{itemize}

\section{Related Work}
BEV representations are widely used to organize multi-camera and multi-sensor driving information in a planner-friendly coordinate frame \cite{palazzi2017lea,philion2020lif,xie2022m$^}. Recent systems such as BEVFormer and BEVFusion use learned BEV features for tasks including object detection and map segmentation \cite{li2025bev,liu2023bev}. BEVFusion is especially relevant here because it fuses camera and lidar information into a shared BEV representation and exposes task-specific outputs that can be reused by external modules.

For data, we rely on nuScenes, a multimodal autonomous driving dataset with diverse urban scenes and rich annotations \cite{caesar2020nusc}. Compared with lidar-only perception baselines such as PointPillars \cite{lang2019poi}, BEVFusion provides both semantic map context and object-level information, which is useful for trajectory assessment.

In planning literature, candidate trajectories are often evaluated through cost functions, reward models, or end-to-end learned evaluators \cite{huang2024dif,liu2021dee,li2025enda}. Our work is closest in spirit to trajectory evaluation methods such as WoTE \cite{li2025enda}, but differs in two important ways. First, our approach is modular and explicitly rule-based rather than end-to-end optimized. Second, we benchmark the ranking output against human preferences to assess whether the induced safety ordering is plausible to human observers.

\section{Method}
\subsection{System Overview}
Camera and lidar data from nuScenes are processed by BEVFusion to obtain BEV object detections and map segmentations. A trajectory evaluation script then accepts an ego start position, an optional target location, and multiple candidate trajectories. Each candidate is simulated on a discrete 200$\times$200 BEV grid and ranked according to safety and task-completion criteria. The resulting visualizations are presented as pairwise choices in the human-preference survey and the responses are compared with the script rankings.

\subsection{Data Extraction and Mapping}
The pipeline uses BEVFusion outputs from camera and lidar inputs originating from nuScenes scenes. Two output modalities are retained for downstream evaluation: object detections and semantic map segmentations. The object detections provide class-aware footprints of relevant traffic participants such as cars, pedestrians, cyclists, buses, and trucks. The segmentation map provides semantic context including drivable area, lane-related markings, pedestrian crossings, walkways, and other road-surface categories.

\begin{figure}[!ht]
    \centering

    \subfloat[BEVFusion detections]{
        \includegraphics[width=0.46\columnwidth]{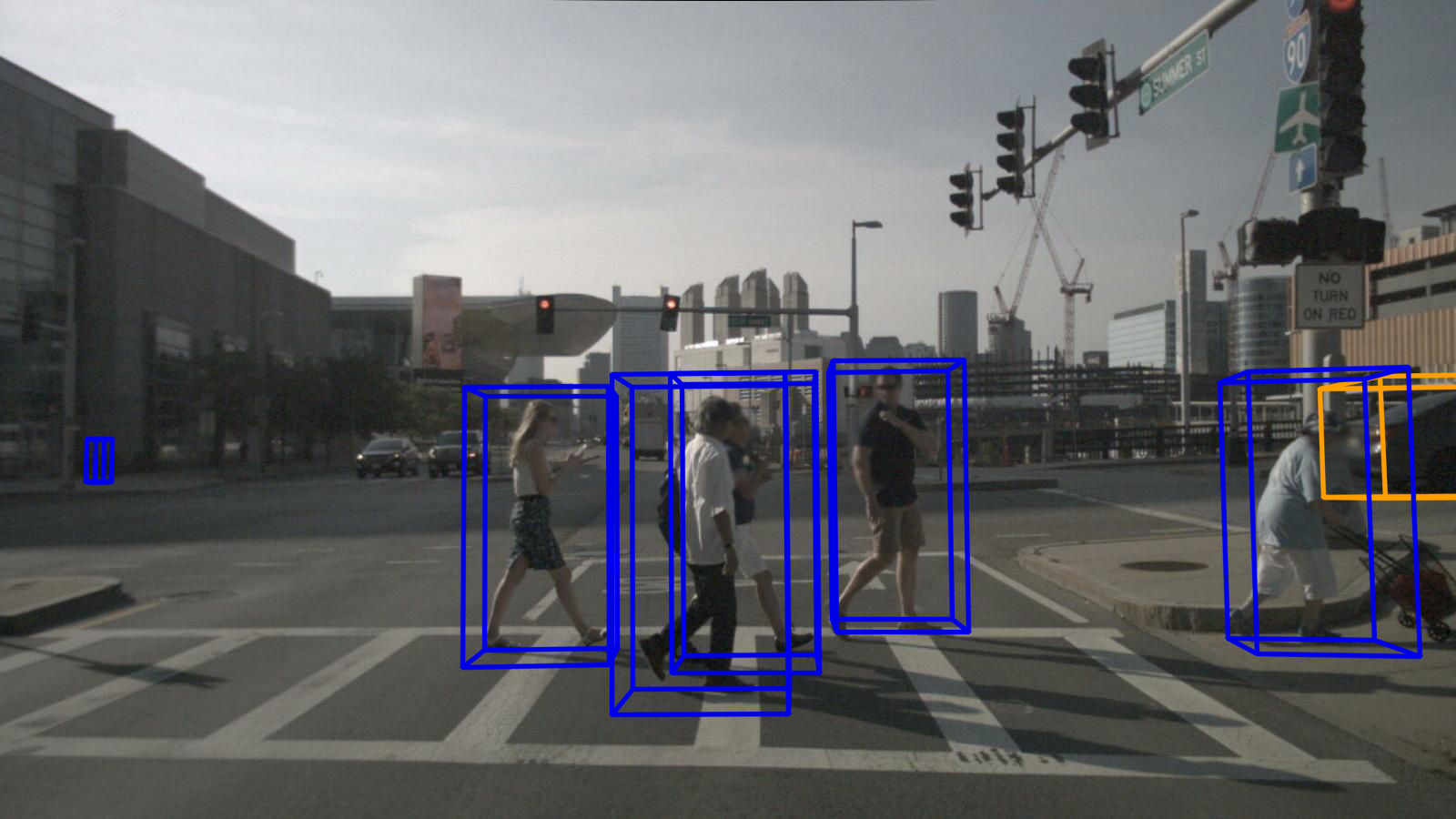}
    }
    \hfill
    \subfloat[Unified BEV representation]{
        \includegraphics[width=0.46\columnwidth]{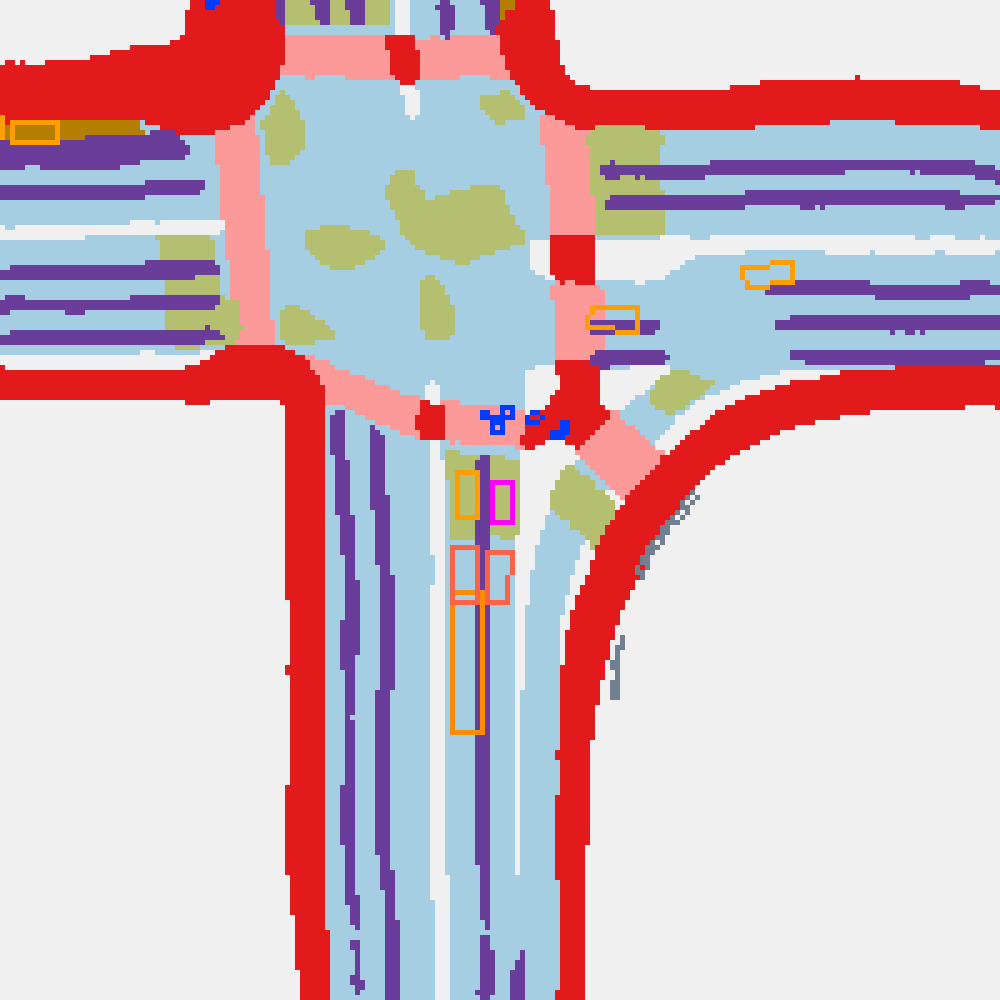}
    }
    \caption{Illustration of the mapping step from BEVFusion object detections to the unified BEV representation used by the trajectory evaluator. For legend see Fig.~\ref{fig:scenarios}. }
    \label{fig:bev_mapping_example}
\end{figure}

Because the output modalities are generated by different prediction heads, a practical integration step is required before safety evaluation can be performed. Our pipeline maps both modalities into a shared 2D BEV canvas so that the ego footprint can be tested against detected objects and semantic classes in a common coordinate frame. Fig.~\ref{fig:bev_mapping_example} illustrates this mapping step using an example from the nuScenes dataset. The left panel shows the BEVFusion object-detection output, while the right panel shows the corresponding unified BEV representation used by the trajectory evaluator.

\subsection{BEV Scenario Representation}
Each scenario combines:
\begin{itemize}
\item BEV map segmentation from BEVFusion,
\item BEV object detections represented as 2D footprints,
\item an ego vehicle pose,
\item an optional target location, and
\item two candidate trajectories.
\end{itemize}

The evaluation deliberately assumes a static environment for non-ego actors. This simplifies the study and isolates the question of whether BEV scene structure alone is sufficient for meaningful safety-oriented trajectory ranking. In this sense, the static BEV representation can be viewed as a simplified world-state abstraction rather than a full dynamic world model.

\subsection{Motion Model and Candidate Encoding}
Candidate ego trajectories are encoded as sequences of discrete moves on the 200$\times$200 BEV grid. The underlying movement logic supports horizontal, vertical, and diagonal transitions, which makes it possible to represent lane following, turning maneuvers, and coarse obstacle avoidance while keeping the evaluator simple. During simulation, the ego footprint is rotated and translated step by step so that each intermediate pose can be scored against the mapped objects and semantic regions.

This discrete setup intentionally favors interpretability over realism. It does not attempt to model steering curvature, velocity profiles, braking dynamics, or control feasibility. Instead, it functions as a mid-level evaluation layer that can compare a small set of candidate plans under a shared BEV scene abstraction.

\subsection{Position Scoring}
At every ego position visited by a trajectory, the script computes an instantaneous ego score. Two safety signals are used:
\begin{enumerate}
\item overlap between the ego footprint and detected objects, interpreted as a collision;
\item overlap between the ego footprint and walkways.
\end{enumerate}

Walkway penalties are weighted by the fraction of ego-occupied tiles belonging to that class. A collision adds a fixed negative penalty. Non-violating positions receive a neutral score. In simplified form, the position score can be written as
\begin{equation}
S_{\mathrm{ego}}=\sum_{c \in C} w_c \cdot r_c + \mathbb{I}_{\mathrm{coll}} \cdot p_{\mathrm{coll}},
\end{equation}
where $r_c$ is the fraction of ego-covered tiles assigned to class $c$, $w_c$ is the class penalty, $\mathbb{I}_{\mathrm{coll}} \in \{0,1\}$ is a binary collision indicator that equals one if the ego footprint overlaps with a detected object and zero otherwise, and $p_{\mathrm{coll}}$ is the collision penalty. In the present implementation, the only nonzero map-class penalty is assigned to walkway tiles. This formulation makes the ranking rule transparent and easy to adapt for additional semantic classes.

\subsection{Trajectory Ranking}
For each candidate trajectory, the script computes three metrics:
\begin{itemize}
\item \textbf{trajectory score}: cumulative sum of the per-step ego scores,
\item \textbf{end distance to target}: Euclidean distance from the trajectory end position to the specified target, and
\item \textbf{distance traveled}: total path length on the grid.
\end{itemize}

Trajectories are ranked lexicographically, prioritizing safety first, followed by smaller target distance and then shorter traveled distance.

This ordering reflects a simple but practical assumption for vehicular safety studies, namely that avoiding collisions and clearly impermissible maneuvers matters more than efficiency, while efficiency only breaks ties among similarly safe candidates.

The cumulative trajectory score is obtained by aggregating the per-step ego scores over the full candidate path, where $T$ denotes the final time step of the prediction horizon:
\begin{equation}
S_{\mathrm{traj}}=\sum_{t=0}^{T} S_{\mathrm{ego}}^{(t)}.
\end{equation}
This penalizes both brief semantic violations and long-duration unsafe trajectories, while longer exposure to risk yields a larger negative total score.

\subsection{Human-Facing Visualization}
To support the survey, the script generates combined BEV visualizations that merge segmentation and object detections and annotate ego state, target location, and candidate trajectories. The resulting image sequences are turned into short side-by-side videos so participants can compare two candidate trajectories under the same scene context.

\section{Experimental Setup}
\input{figs/scenarios_figure.tex}
\subsection{Scenario Design}
We created seven scenario pairs from BEVFusion outputs generated on nuScenes scenes. 
Each scenario contains two candidate trajectories and was designed to isolate one dominant 
comparison factor:

\begin{itemize}
\item \textbf{Scenario~1: Vehicle collision} (cf. Fig.~\ref{fig:scenario-vehicle}). 
One trajectory collides with a vehicle, whereas the alternative avoids the obstacle.

\item \textbf{Scenario~2: Walkway intrusion} (cf. Fig.~\ref{fig:scenario-walkway}). 
One trajectory drives over a walkway, whereas the alternative remains on the regular 
driving area.

\item \textbf{Scenario~3: Path efficiency} (cf. Fig.~\ref{fig:scenario-route-length}). 
Both trajectories are safe and reach the target, but one trajectory follows a shorter route.

\item \textbf{Scenario~4: Target proximity} (cf. Fig.~\ref{fig:scenario-target}). 
One trajectory reaches the target location, whereas the alternative stops before reaching it, leading to a larger remaining end distance.

\item \textbf{Scenario~5: Final target distance} (cf. Fig.~\ref{fig:scenario-final-distance}). 
Both trajectories remain safe, but they end at different distances from the target location.

\item \textbf{Scenario~6: Pedestrian collision} (cf. Fig.~\ref{fig:scenario-pedestrian}). 
One trajectory conflicts with a pedestrian, whereas the alternative avoids the conflict.

\item \textbf{Scenario~7: Cyclist collision} (cf. Fig.~\ref{fig:scenario-cyclist}). 
One trajectory conflicts with a cyclist, whereas the alternative avoids the conflict.
\end{itemize}

The scenarios are intentionally constrained and interpretable rather than statistically 
exhaustive. Their purpose is to test whether the proposed ranking logic reflects plausible 
human judgments across common safety and task-completion trade-offs. Representative BEV 
visualizations for all seven scenarios are shown in Fig.~\ref{fig:scenarios}.

Across the seven comparisons, the evaluator is exposed to both hard safety violations and softer trade-offs. This is important because a method that only succeeds on obvious collision scenarios would have limited practical value. By including target-proximity and path-efficiency cases, the study also probes whether the ranking remains plausible when safety penalties are absent or equal between candidates.

\subsection{Human Preference Survey}
\begin{figure}[!ht]
    \vspace{2mm}
\centering
\includegraphics[width=\columnwidth]{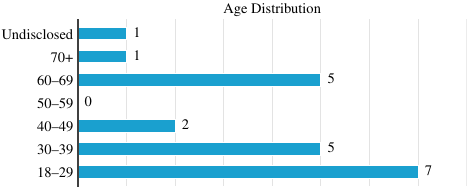}
\par\vspace{0.12em}
\includegraphics[width=\columnwidth]{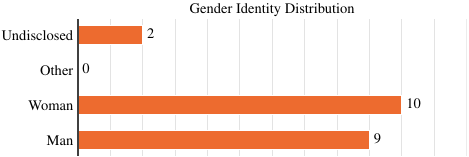}
\caption{Participant demographics for the human preference survey. Top: age distribution. Bottom: gender identity distribution.}
\label{fig:demographics}
\end{figure}

Twenty-one participants completed a survey in which each scenario was presented as a pairwise comparison between two trajectory videos. Participant demographics, including age and gender identity distributions, are summarized in Fig.~\ref{fig:demographics}. Participants were introduced to the BEV legend, traffic participants, road elements, and the general decision context, but were encouraged to use their own judgment when choosing the preferable trajectory.

The survey was designed as a benchmark for human preferences rather than as a statistically representative population study. Its primary purpose was to test whether the rankings induced by the script appear reasonable to human observers under the same informational constraints. Participants therefore viewed the same simplified BEV visualizations that the system itself was derived from, rather than richer sensor footage.

\section{Results}
Table~\ref{tab:script-results} summarizes the script-derived rankings. In scenarios with collisions or walkway violations, the safety-first rule strongly separates the alternatives. In target-driven cases without explicit safety violations, the ranking falls back to end distance to the target and path efficiency.

\begin{table}[!th]
  \vspace{2mm}
\caption{Script-generated trajectory metrics and rankings for the seven survey scenarios.}
\label{tab:script-results}
\centering
\begin{tabular}{|c|c|c|c|c|c|}
\hline
Scenario & Traj. & Score & Target Dist. & Path Len. & Rank \\
\hline
1 & 1 & 0 & 0 & 60 & 1 \\
1 & 2 & -160 & 0 & 60 & 2 \\
\hline
2 & 1 & -71 & 0 & 69 & 2 \\
2 & 2 & 0 & 0 & 75 & 1 \\
\hline
3 & 1 & 0 & 0 & 97 & 2 \\
3 & 2 & 0 & 0 & 85 & 1 \\
\hline
4 & 1 & 0 & 0 & 87 & 1 \\
4 & 2 & 0 & 8 & 84 & 2 \\
\hline
5 & 1 & 0 & 29 & 95 & 2 \\
5 & 2 & 0 & 14 & 109 & 1 \\
\hline
6 & 1 & -100 & N/A & 61 & 2 \\
6 & 2 & 0 & N/A & 76 & 1 \\
\hline
7 & 1 & -120 & N/A & 79 & 2 \\
7 & 2 & 0 & N/A & 72 & 1 \\
\hline
\end{tabular}%
\end{table}

Table~\ref{tab:agreement-results} compares the preferred script trajectory with human choices. Agreement is perfect in the first two scenarios and remains high in the pedestrian and route-efficiency cases. The weakest alignment occurs in Scenarios 4, 5, and 7, where the decision involves target progress, maneuver simplicity, or less salient safety cues rather than an obvious violation.

\begin{table}[!th]
    \vspace{2mm}
\caption{Agreement between the script-preferred trajectory and human survey responses.}
\label{tab:agreement-results}
\centering
\begin{tabular}{|c|c|c|}
\hline
Scenario & Matching Votes & Agreement \\
\hline
1 & 21/21 & 100.0\% \\
2 & 21/21 & 100.0\% \\
3 & 19/21 & 90.5\% \\
4 & 15/21 & 71.4\% \\
5 & 15/21 & 71.4\% \\
6 & 20/21 & 95.2\% \\
7 & 13/20 & 65.0\% \\
\hline
All & 124/146 & 84.9\% \\
\hline
\end{tabular}
\end{table}

\subsection{Baseline and Ranking Ablation}
To assess whether the specific ranking rule matters, we ranked the metrics in
Table~\ref{tab:script-results} again using individual criteria, an equal-weight sum of
per-scenario normalized criteria, and alternative lexicographic priorities.
Let $S$, $D$, and $L$ denote trajectory score (higher is better), end distance
to target (lower is better), and path length (lower is better), respectively.
A match is a vote in which the rule selects the same trajectory as the
participant; agreement is matches divided by the 146 total votes.
When an ablated rule ties a pair, we assign each trajectory probability $0.5$,
so tied scenarios contribute half their votes in expectation. The binary
random-choice baseline is 50\%.

\begin{table}[!th]
\caption{Baseline and ranking ablation on the same 146 survey votes.}
\label{tab:ranking-ablation}
\centering
\footnotesize
\begin{tabular}{|l|c|c|c|}
\hline
Ranking Rule & Ties & Expected Matches & Agreement \\
\hline
Random binary choice & -- & 73.0/146 & 50.0\% \\
$S$ only & 3 & 106.5/146  & 72.9\% \\
$D$ only & 5 & 82.0/146  & 56.2\% \\
$L$ only & 1 & 55.5/146  & 38.0\% \\
Equal-weight sum & 4 & 95.0/146  & 65.1\% \\
$S>D>L$ (proposed) & 0 & 124/146  & \textbf{84.9\%} \\
$S>L>D$ & 0 & 106/146  & 72.6\% \\
$D>S>L$ & 0 & 124/146  & \textbf{84.9\%} \\
$L>D>S$ & 0 & 66/146  & 45.2\% \\
\hline
\end{tabular}
\end{table}

The proposed $S>D>L$ ordering achieves the highest agreement at 84.9\%
(124 matches), exceeding the random baseline and every single-criterion rule:
$S$ alone 72.9\%, $D$ alone 56.2\%, and $L$ alone 38.0\%, which is below the baseline,
confirming that path length alone is a poor proxy for human preference. The equal-weight sum reaches 65.1\% because conflicting criterion preferences produce four ties that the lexicographic ordering resolves.
Swapping the lower-priority criteria ($S>L>D$) drops to 72.6\%, as the $D$/$L$ order affects Scenarios~4 and~5. The $D>S>L$ variant also achieves 84.9\%. $S$ and $D$ never favor different trajectories; when one distinguishes between them, the other is tied. This invariance shows the result is robust to the relative order of the two criteria; the critical choice is that both precede path length.

The 84.9\% agreement suggests that a simple ranking strategy can approximate human safety preferences in constrained BEV choice tasks, though lower agreement in scenarios 4, 5, and~7 highlights the limits of static BEV-only reasoning.

\subsection{Per-Scenario Observations}
The strongest agreement occurs in scenarios with visually salient and semantically unambiguous violations. Participants unanimously reject the trajectory that collides with a car in scenario 1 and the trajectory that drives onto a walkway in scenario 2. This indicates that the evaluator's safety-first bias is well aligned with human expectations when the unsafe maneuver is obvious from the BEV presentation.

Scenario 3 also yields high agreement because both trajectories are safe and reach the target, leaving route efficiency as the key differentiator. Here, the shorter path is selected by both the script and the large majority of participants. This is encouraging because it shows that the tie-breaking logic is not only mathematically simple but also behaviorally plausible.

The lower-agreement cases provide additional insight. In scenario 4, nearly 30\% of participants prefer the trajectory that does not quite reach the target, which suggests that some viewers favor the visually simpler straight path over a lane-changing maneuver when the remaining target distance appears modest. Scenario 5 exhibits a similar 15--6 split, but in the opposite geometric setting: both trajectories miss the target, and some participants appear to prefer the route with more immediate straight progress even though it ends farther away overall. These two cases indicate that humans may implicitly weigh directness and maneuver simplicity alongside the explicit ranking criteria.

Scenario 7 is particularly notable because 7 of 20 respondents select the trajectory associated with a cyclist collision, while one participant provides no response. Given the unanimous rejection of the vehicle collision in scenario 1 and the near-unanimous rejection of the pedestrian collision in scenario 6, a plausible explanation is not that participants consciously accept the cyclist collision, but that the cyclist is harder to perceive in the simplified BEV rendering. This suggests that part of the disagreement may stem from visualization fidelity rather than from the ranking logic alone.

\ApplyICVESBalanceAdjustment   

\section{Discussion}
The main value of this work is to examine how much safety-relevant trajectory preference information can be recovered from static BEV perception outputs using transparent ranking rules. Because the ranking logic is explicit, the method can support rapid safety screening, validation of BEV scenario interpretations with human input, dataset curation for trajectory preference studies, and explainable baselines for more advanced learned evaluators.

The approach is computationally lightweight compared with planning models trained end to end, making it suitable for prototyping and offline evaluation when the goal is to inspect safety trade-offs rather than optimize a full autonomy stack.

The results also speak to the role of BEV abstractions in safety validation. Even though the evaluator only consumes static BEV outputs and coarse semantic classes, it captures a large fraction of human pairwise preferences, suggesting that interpretable heuristics at the BEV level can serve as useful intermediate artifacts for analyzing behavior from perception to planning. 

However, the ranking depends not only on the scoring rule but also on how clearly the BEV representation shows the relevant scene information: when safety-relevant objects and map regions are clearly represented, simple rules identify plausible preferences, but when important cues are subtle or difficult to see, disagreement may result from limitations of the representation or visualization rather than from the ranking rule alone.

\section{Limitations and Future Work}
The current study is limited to static BEV scenes and therefore ignores temporal interactions as well as context-dependent driving cues such as traffic control, speed constraints, and lane directionality. In addition, the 200$\times$200 segmentation grid is coarse, and some disagreements may stem from the human-facing visualization rather than the ranking rule itself.

The evaluation also treats BEVFusion outputs as fixed inputs and does not analyze how perception uncertainty affects the ranking. Missed detections, false positives, inaccurate object footprints, or noisy segmentation boundaries could alter collision and semantic-violation scores. Moreover, the human benchmark is based on a small, manually curated set of pairwise comparisons and should therefore be interpreted as evidence of plausibility rather than broad behavioral validity.

Future work should extend the evaluator toward dynamic and semantically richer scene representations, for example by assessing candidate trajectories against predicted future BEV states. Additional scenarios should place safety violations in direct conflict with target progress and path efficiency so that penalty magnitudes and the full priority ordering can be identified. Interpretable ranking methods could also support automated trajectory curation in downstream dataset-construction pipelines\icvescitebagzel

\section{Conclusion}
This paper investigated safety-oriented trajectory ranking from static BEV perception outputs. Using BEVFusion object detections and map segmentations from nuScenes data, candidate ego trajectories were ranked with transparent criteria covering collisions, walkway violations, end distance to the target, and path efficiency. In seven controlled scenario pairs, the ranking agreed with human preferences in 84.9\% of the survey responses.

The results indicate that interpretable rule-based reasoning over BEV outputs can capture important aspects of human safety judgment in constrained trajectory choice tasks. At the same time, disagreements in more ambiguous scenarios underline the limitations of static BEV reasoning, particularly when comfort, intent, or subtle traffic participants influence human decisions. Future work should extend this evaluation toward dynamic BEV representations and semantically richer trajectory assessment with predictive world models.

\section*{Acknowledgment}

\ificvesanonymous
The authors acknowledge support from ... (Anonymized due to double-blind review).
\else
The authors acknowledge support from BAAINBw and dtec.bw -- Digitalization and Technology Research Center of the Bundeswehr [project MORE]. dtec.bw is funded by the European Union -- NextGenerationEU.
\fi

\bibliographystyle{IEEEtran}
\bibliography{lepo.bib, et_al}

%% file: figs/teaser.tex
\definecolor{lightblue}{RGB}{225,240,248}
\definecolor{lightgreen}{RGB}{225,240,225}
\definecolor{softgray}{RGB}{245,245,245}
\definecolor{lightyellow}{RGB}{252,248,224}

\begin{tikzpicture}[
    font=\sffamily,
    imagecard/.style={
        rectangle,
        rounded corners=4pt,
        draw=black,
        line width=0.8pt,
        fill=white,
        align=center,
        inner sep=4pt
    },
    card/.style={
        rectangle,
        rounded corners=4pt,
        draw=black,
        line width=0.8pt,
        fill=white,
        align=center,
        inner sep=5pt,
        font=\sffamily\scriptsize
    },
    groupcard/.style={
        rectangle,
        rounded corners=5pt,
        draw=black,
        line width=0.8pt,
        fill=lightblue,
        inner sep=5pt
    },
    arrow/.style={
        -{Latex[length=2.4mm,width=1.8mm]},
        line width=0.8pt,
        shorten >=2pt,
        shorten <=2pt,
        line cap=round,
        line join=round
    }
]

\node[imagecard] (det) {
    \includegraphics[width=0.34\columnwidth]{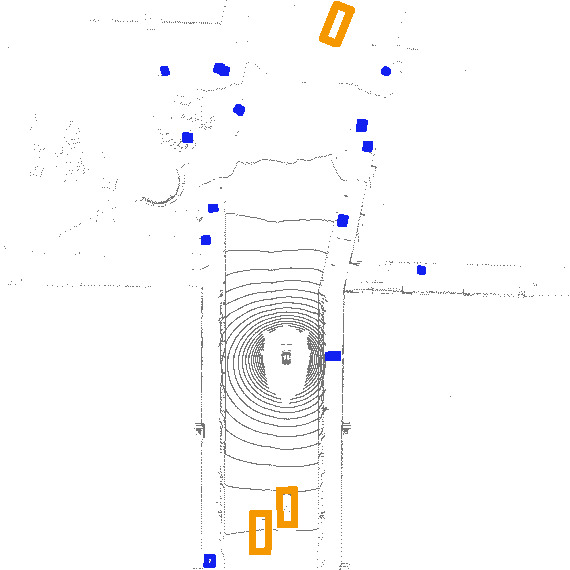}\\[-1pt]
    {\bfseries\scriptsize Object Detections}
};

\node[imagecard, right=0.35cm of det] (seg) {
    \includegraphics[width=0.34\columnwidth]{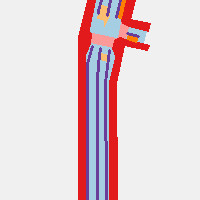}\\[-1pt]
    {\bfseries\scriptsize Map Segmentation}
};

\begin{scope}[on background layer]
\node[
    groupcard,
    fit=(det)(seg)
] (bevoutputs) {};
\end{scope}

\node[
    font=\sffamily\scriptsize,
    above=0.06cm of bevoutputs
] {\textbf{BEVFusion Outputs}};

\node[
    imagecard,
    below=0.55cm of bevoutputs.south west,
    anchor=north west
] (combined) {
    \includegraphics[width=0.47\columnwidth]{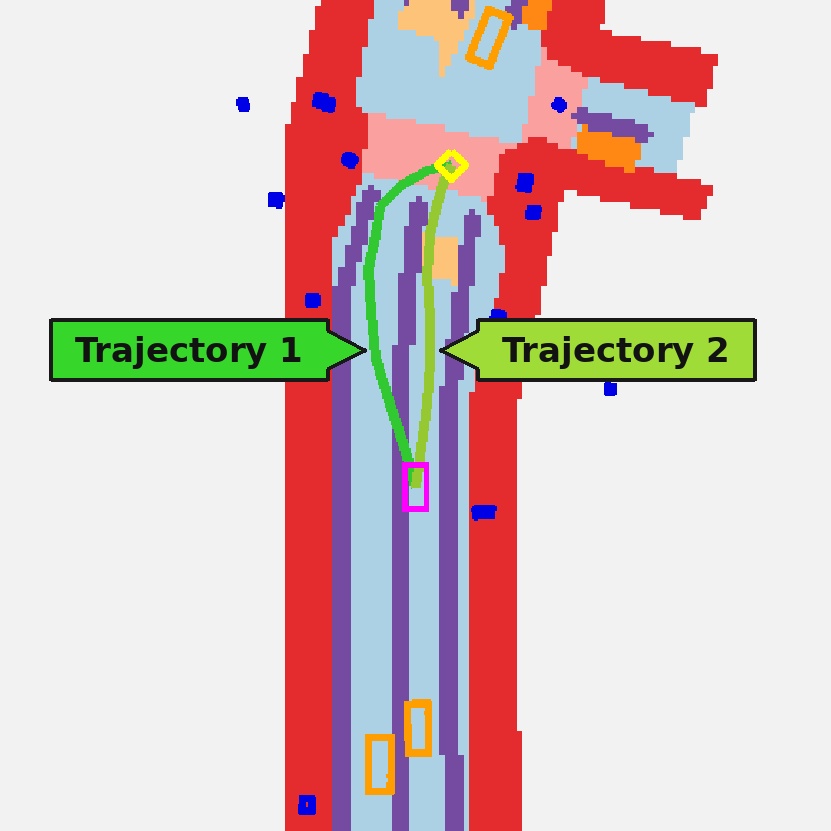}\\[-1pt]
    {\bfseries\scriptsize Unified BEV}
};

\node[
    card,
    fill=lightgreen,
    right=0.60cm of combined,
    yshift=0.95cm,
    minimum width=0.28\columnwidth
] (score) {
    \textbf{Trajectory Scoring}\\[-1pt]
    \tiny cumulative per-step ego penalties\\[2pt]
    $S_{\mathrm{traj}}=\sum_t S_{\mathrm{ego}}^{(t)}$\\[2pt]
    \tiny collision penalty\\
    \tiny walkway penalty
};

\node[
    card,
    fill=lightyellow,
    below=0.50cm of score,
    minimum width=0.28\columnwidth
] (rank) {
    \textbf{Lexicographic Ranking}\\[3pt]
    \tiny
    \begin{tabular}{@{}r@{\hspace{0.35em}}l@{}}
    1. & highest trajectory score\\
    2. & smallest end distance to target\\
    3. & shortest path length
    \end{tabular}
};


\draw[arrow]
    ([xshift=-0.15cm]bevoutputs.south)
    -- ++(0,-0.16)
    -| 
    ([xshift=-0.18cm]combined.north);

\draw[arrow]
    ([yshift=0.42cm]combined.east)
    -- ++(0.18,0)
    |- 
    ([yshift=0.18cm]score.west);

\draw[arrow]
    (score.south)
    --
    (rank.north);

\end{tikzpicture}

%% file: figs/scenarios_figure.tex
\input{figs/legend.tex}

\newcommand{\scenariofigwidth}{0.319\textwidth}
\newcommand{\scenariosevenfigwidth}{0.34\textwidth}
\newcommand{\legendfigwidth}{0.45\textwidth}
\newcommand{\scenariofigsep}{0.3em}
\newcommand{\hspacelegend}{0em}

\newcommand{\scenarioimage}[5]{%
  \begin{tikzpicture}
    \node[anchor=south west, inner sep=0] (img) at (0,0)
      {\includegraphics[width=#1]{#2}};
    \begin{scope}[x={(img.south east)}, y={(img.north west)}]
      #5
      \node[
        fill=white,
        draw=black,
        rounded corners=1pt,
        inner sep=1.4pt,
        font=\footnotesize\bfseries
      ] at #3 {T1};
      \node[
        fill=white,
        draw=black,
        rounded corners=1pt,
        inner sep=1.4pt,
        font=\footnotesize\bfseries
      ] at #4 {T2};
    \end{scope}
  \end{tikzpicture}%
}

\newcommand{\scenariohazard}[4]{%
  \node[
    circle,
    draw=black,
    double=white,
    double distance=0.45pt,
    line width=0.7pt,
    minimum size=11pt,
    inner sep=0pt
  ] (#1-marker) at #2 {};
  \node[
    fill=white,
    fill opacity=0.92,
    text opacity=1,
    draw=black,
    rounded corners=0.8pt,
    inner sep=1pt,
    font=\scriptsize\bfseries
  ] (#1-label) at #3 {#4};
  \draw[-{Latex[length=1.2mm,width=0.9mm]}, line width=0.55pt]
    (#1-label) -- (#1-marker);
}

\begin{figure*}[htbp]
\centering

\subfloat[Scenario~1: Vehicle collision.]{
  \scenarioimage
    {\scenariofigwidth}
    {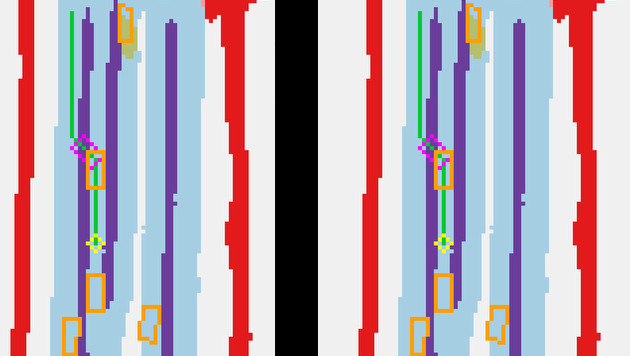}
    {(0.39,0.93)}
    {(0.95,0.93)}
    {\fill[white] (0.432,0) rectangle (0.508,1);}
  \label{fig:scenario-vehicle}
}
\hfill
\subfloat[Scenario~2: Walkway intrusion.]{
  \scenarioimage
    {\scenariofigwidth}
    {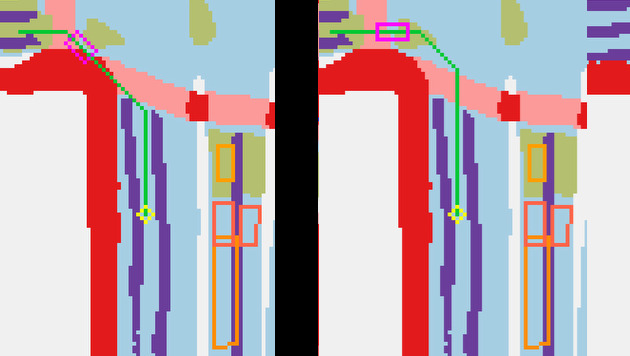}
    {(0.39,0.93)}
    {(0.95,0.93)}
    {\fill[white] (0.432,0) rectangle (0.508,1);}
  \label{fig:scenario-walkway}
}
\hfill
\subfloat[Scenario~3: Path efficiency.]{
  \scenarioimage
    {\scenariofigwidth}
    {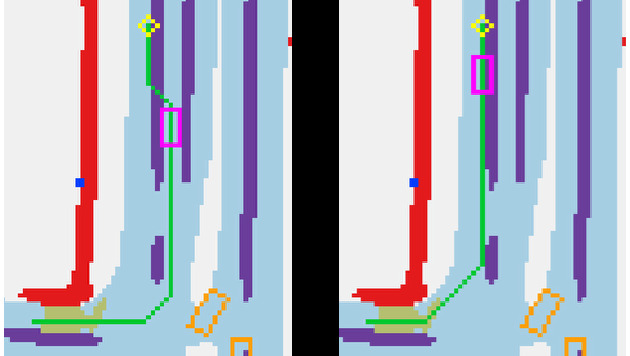}
    {(0.42,0.93)}
    {(0.95,0.93)}
    {\fill[white] (0.458,0) rectangle (0.543,1);}
  \label{fig:scenario-route-length}
}

\vspace{\scenariofigsep}

\subfloat[Scenario~4: Target proximity.]{
  \scenarioimage
    {\scenariofigwidth}
    {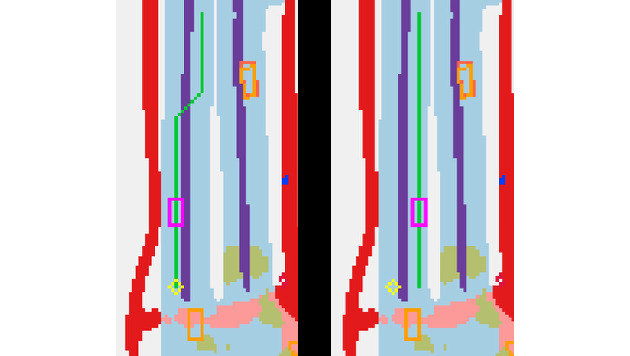}
    {(0.43,0.93)}
    {(0.779,0.93)}
    {\fill[white] (0.468,0) rectangle (0.530,1);}
  \label{fig:scenario-target}
}
\hfill
\subfloat[Scenario~5: Final target distance.]{
  \scenarioimage
    {\scenariofigwidth}
    {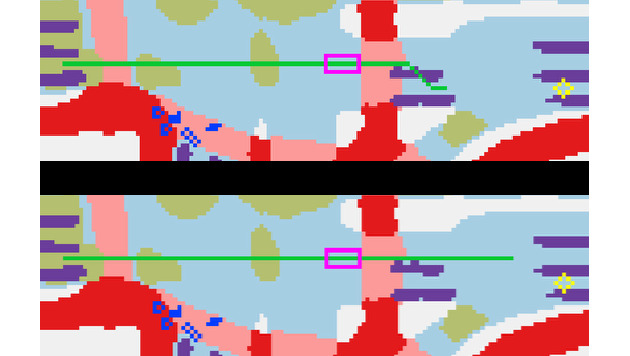}
    {(0.89,0.93)}
    {(0.89,0.39)}
    {\fill[white] (0,0.447) rectangle (1,0.553);}
  \label{fig:scenario-final-distance}
}
\hfill
\subfloat[Scenario~6: Pedestrian collision.]{
  \scenarioimage
    {\scenariofigwidth}
    {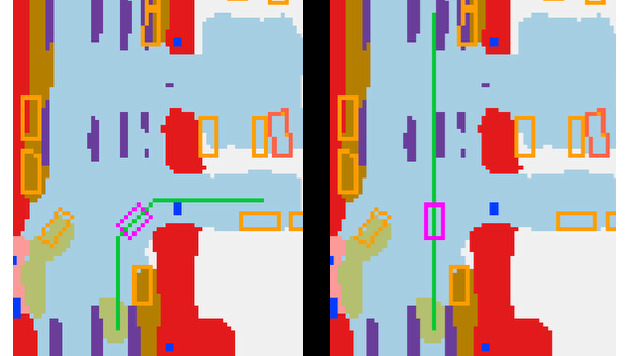}
    {(0.435,0.93)}
    {(0.935,0.93)}
    {%
      \fill[white] (0.476,0) rectangle (0.528,1);
      \scenariohazard{s6-t2}{(0.7845,0.413)}{(0.750,0.570)}{ped.}
      
    }
  \label{fig:scenario-pedestrian}
}

\vspace{\scenariofigsep}

\makebox[\textwidth][c]{%
  \subfloat[Scenario~7: Cyclist collision.]{
    \scenarioimage
      {\scenariosevenfigwidth}
      {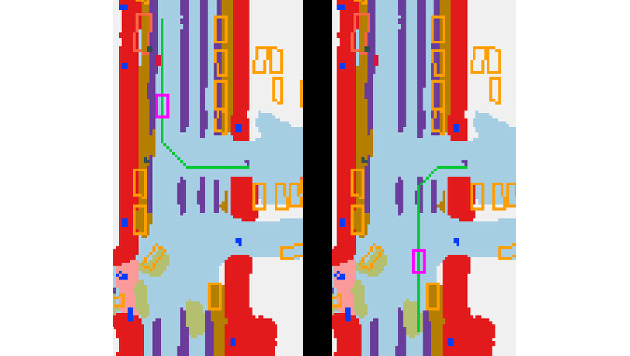}
      {(0.44,0.95)}
      {(0.78,0.95)}
      {%
        \fill[white] (0.476,0) rectangle (0.531,1);
        \scenariohazard{s7-t2}{(0.596,0.831)}{(0.6305,0.674)}{cyclist}
      }
    \label{fig:scenario-cyclist}
  }
  \hspace{\hspacelegend}
  \begin{minipage}[b]{\legendfigwidth}
    \centering
    \scenariolegend[0.75cm]
  \end{minipage}
}

\caption{Representative BEV visualizations for all seven survey scenarios, where T1 and T2 denote trajectory 1 and trajectory 2.
}
\label{fig:scenarios}
\end{figure*}

%% file: figs/legend.tex
\definecolor{carcolor}{RGB}{255,165,5}
\definecolor{truckcolor}{RGB}{250,107,75}
\definecolor{bicyclecolor}{RGB}{246,19,65}
\definecolor{pedcolor}{RGB}{4,55,255}
\definecolor{conecolor}{RGB}{56,86,86}
\definecolor{egocolor}{RGB}{255,0,255}
\definecolor{trajectorycolor}{RGB}{0,199,48}

\definecolor{drivablecolor}{RGB}{165,206,226}
\definecolor{crossingcolor}{RGB}{253,153,151}
\definecolor{walkwaycolor}{RGB}{255,25,27}
\definecolor{dividercolor}{RGB}{107,62,155}
\definecolor{stoplinecolor}{RGB}{189,192,110}
\definecolor{carparkcolor}{RGB}{178,127,0}
\definecolor{nodatacolor}{RGB}{241,241,241}

\newcommand{\legenditem}[2]{%
  \tikz[baseline={(0,0.01)}]{
    \draw[black, line width=0.5pt, fill=#1] (0,0) rectangle (0.18,0.18);
  }~#2%
}

\newcommand{\scenariolegend}[1][0pt]{%
  \raisebox{#1}{%
    \begin{tikzpicture}
      \node[
        draw=black!45,
        fill=white,
        rounded corners=1pt,
        inner xsep=4pt,
        inner ysep=2.5pt,
        line width=0.25pt
      ] {
        \footnotesize
        \begin{tabular}{@{}ll@{\hspace{0.5em}}ll@{}}
          \multicolumn{2}{c}{\bfseries Objects} &
          \multicolumn{2}{c}{\bfseries Map Segments} \\[1pt]
          \legenditem{carcolor}{car} &
          \legenditem{truckcolor}{truck} &
          \legenditem{drivablecolor}{drivable area} &
          \legenditem{crossingcolor}{ped.\ crossing} \\
          \legenditem{bicyclecolor}{bicycle} &
          \legenditem{pedcolor}{pedestrian} &
          \legenditem{walkwaycolor}{walkway} &
          \legenditem{dividercolor}{divider} \\
          \legenditem{conecolor}{traffic cone} &
          \legenditem{egocolor}{EGO} &
          \legenditem{stoplinecolor}{stop line} &
          \legenditem{carparkcolor}{carpark area} \\
          \multicolumn{2}{c}{} &
          \legenditem{trajectorycolor}{trajectory} &
          \legenditem{nodatacolor}{no data}           
        \end{tabular}
      };
    \end{tikzpicture}%
  }%
}